\documentclass[preprint,12pt]{elsarticle}

\usepackage{graphicx}
\usepackage{hyperref}
\usepackage{graphics}
\usepackage{color}
\usepackage{multirow}
\usepackage{geometry}
\usepackage{array}
\usepackage{adjustbox}
\usepackage{rotating}

\usepackage{caption}
\usepackage{subcaption}
\usepackage{times}
\usepackage{latexsym}
\usepackage{xcolor}
\usepackage{longtable}

\usepackage{microtype}
\usepackage{amsfonts}
\usepackage{amsmath,bm}
\usepackage{amssymb}

\usepackage{mathptmx}
\usepackage{soul}\setuldepth{article}
\usepackage{algorithmic}
\usepackage{xcolor}
\usepackage[title]{appendix}
\usepackage[utf8]{vietnam}
\usepackage[british]{babel}
\usepackage[utf8]{inputenc}
\usepackage{hyperref}

\newcolumntype{P}[1]{>{\centering\arraybackslash}p{#1}}

\usepackage{lineno,hyperref}
\modulolinenumbers[5]

\newcommand{\mc}{\mathcal}

\def\1{\bm{1}}
\def\vx{{\bm{x}}}
\def\ve{{\bm{e}}}
\def\vm{{\bm{m}}}

\journal{Information Sciences}

\begin{document}

\begin{frontmatter}

\title{Learning for Amalgamation: A  Multi-source Transfer Learning Framework  for Sentiment Classification}

\author[seventhaddress,eighthaddress,ninthaddress]{Cuong V. Nguyen}
\author[seventhaddress,eighthaddress,ninthaddress]{Khiem H. Le}
\author[seventhaddress,eighthaddress,ninthaddress]{Anh M. Tran}
\author[fourthaddress]{Quang H. Pham}
\author[seventhaddress,eighthaddress,ninthaddress]{Binh T. Nguyen\corref{correspondingauthor}}
\cortext[correspondingauthor]{Corresponding author}
\ead{ngtbinh@hcmus.edu.vn}
\address[fourthaddress]{Singapore Management University}
\address[seventhaddress]{AISIA Research Lab, Ho Chi Minh City, Vietnam}
\address[eighthaddress]{Department of Computer Science, University of Science, Ho Chi Minh City, Vietnam}
\address[ninthaddress]{Vietnam National University, Ho Chi Minh City, Vietnam}

\begin{abstract}

Transfer learning plays an essential role in Deep Learning, which can remarkably improve the performance of the target domain, whose training data is not sufficient. Our work explores beyond the common practice of transfer learning with a single pre-trained model. We focus on the task of Vietnamese sentiment classification and propose LIFA, a framework to learn a unified embedding from several pre-trained models. 
We further propose two more LIFA variants that encourage the pre-trained models to either cooperate or compete with one another. Studying these variants sheds light on the success of LIFA by showing that sharing knowledge among the models is more beneficial for transfer learning. Moreover, we construct the AISIA-VN-Review-F dataset, the first large-scale Vietnamese sentiment classification database. We conduct extensive experiments on the AISIA-VN-Review-F and existing benchmarks to demonstrate the efficacy of LIFA compared to other techniques. To contribute to the Vietnamese NLP research, we publish our source code and datasets to the research community upon acceptance.

\end{abstract}


\end{frontmatter}


\section{Introduction}
\label{sec:introduction}
Sentiment analysis has been extensively studied for the last two decades and has had a lot of practical applications in natural language processing (NLP), data mining (DM), information retrieval (IR), social networks, and e-commerce \cite{WANG201477,10.1007/978-3-319-77712-2_12}.
With the rise of deep learning, there has been tremendous success in sentiment classification for popular languages such as English and Chinese \cite{zhang2018deep}. Besides the expressive power of deep neural networks, this success is also attributed to the transfer learning approach with high-quality pre-trained models such as Word2Vec \cite{Word2Vec}, BERTs \cite{BERT, PhoBERT}, and GPTs \cite{GPT2}. 
Given the ubiquitous of pre-trained models, one might face the \emph{multi-source transfer learning} problem where it is difficult to choose the appropriate pre-trained model for a task at hand. Moreover, it could be more beneficial to take advantage of several pre-trained models based on different architectures and trained on a diverse corpus.

In the multi-source transfer learning problem, most existing methods assume having access to a set of source datasets and aim to transfer the source knowledge to perform well on a single target dataset. 
Various strategies have been developed under this setting and have shown promising results across a wide range of applications \cite{Long_2015,Gupta_2008}. However, having access to many source datasets might not be a realistic assumption, and can even be prohibited in real-world scenarios. In practice, it is more common to obtain and use a pre-trained model per source dataset, while the source data are kept away from access due to privacy concerns. Therefore, our work focuses on the multi-source transfer learning scenario where one only has access to the pre-trained models rather than the source data. This setting has recently gained much interest~\cite{NEURIPS2019_6048ff4e}, but not yet widely explored in the NLP domain.

In this paper, we propose LIFA (\textbf{L}earn\textbf{I}ng \textbf{F}or \textbf{A}lmagamation), a  learning framework to combine various pre-trained models from one source dataset into a unified embedding that can perform better than its components in the target task.  Motivated from Mixture of Experts (MOE) \cite{MoE}, LIFA introduces an additional gating layer trained to combine existing embeddings and produce the final embedding for classification. As a result, without having access to the training data, our proposed LIFA takes advantage of rich-knowledge sources and allows our sentiment classification to leverage features dynamically and selectively from each source through a probabilistic mixture of expert mechanisms. Notably, it helps tackle the shortcomings of existing algorithms while only acquiring a small amount of data for training and boosting performance remarkably. 

We conduct extensive experiments on four different datasets: two Vietnamese datasets, including one public dataset of AIVIVN and our large-scale dataset collected from Vietnamese e-commerce websites (namely AISIA-VN-Review-F Dataset), and two multi-domains English benchmark datasets, including Multi-Domain Dataset and Amazon Reviews Dataset. We consider three base transferring sources for each classification problem or Vietnamese or English. For Vietnamese classification, we employ the sources of Fasttext \cite{joulin2016fasttext}, BERT \cite{BERT} and PhoBERT \cite{PhoBERT}. For English classification, we use the sources of FastText \cite{joulin2016fasttext}, BERT \cite{BERT}, XLM \cite{XLM}. Our LIFA variants of SIGMOID, WTA, COOP consider all three sources and learn how to combine them. The results show that our LIFA-SIGMOID consistently outperforms other approaches that transfer only from a single source and show a better performance than a traditional ensembling method of concatenation.
In summary, our work makes the following contributions:
\begin{enumerate}[(a)]
    \item We propose LIFA, a novel framework for transfer learning using multiple pre-trained sources with different embedding sizes. We also consider and compare different variants of LIFA, including LIFA-SIGMOID, LIFA-WTA, and LIFA-COOP.
    \item Through extensive experiments, we demonstrate the efficacy of our proposed LIFA compared to other existing techniques. Meanwhile, LIFA-SIGMOID shows the best performance.
    \item We construct the AISIA-VN-Review-F Dataset consisting of over 450K reviewing comments that we manually labeled. We will publish the AISIA-VN-Review-F Dataset, both the raw and post-processed versions, and LIFA's implementation to facilitate the research community in Vietnamese sentiment analysis.
\end{enumerate}

The rest of this paper is organized as follows. Sections \ref{sec:introduction} and \ref{sec:related_work} formulate the main problem and provide an overview of the literature. In Section \ref{sec:method}, we present 
our proposed LIFA framework with different variants (LIFA-SIGMOID, LIFA-WTA, LIFA-COOP). In Section \ref{sec:exp}, we introduce the AISIA-VN-Review-F dataset and conduct extensive experiments to validate the efficacy of LIFA compared to existing techniques. Finally, we conclude this work in Section~\ref{sec:conclusion}.

\section{Related Work}
\label{sec:related_work}
\subsection{Overview} 
There have many studies investigated the problem of sentiment classification. Traditionally, the methods usually employ the word embedding Word2Vec \cite{Word2Vec} or Fasttext \cite{joulin2016fasttext} to transform sentences to vectors, design raw architectures, and train from scratch such as deep character-level CNNs \cite{DeepCharCNN}, shallow word-level \cite{ShallowWordCNN}, recurrent networks \cite{LSTM}, combination of convolutional and recurrent networks \cite{RCNN}, or residual-based networks \cite{DPCNN}. However, these methods' drawbacks come from the requirements to design and test with many raw architectures and train on a very large-scale dataset to guarantee a good performance. With the rising of transfer learning, it has become increasingly common to utilize pre-trained models and finetune on a downstream task. The pre-trained models are usually trained on very large-scale datasets and heavily designed with a complex architecture of million or billion parameters. This approach has been successfully applied and obtained state-of-the-art results in many of the most common NLP benchmarks but mainly limited to the English language such as BERT \cite{BERT}, XLNET \cite{XLNet}, XLM \cite{XLM}, or UniLM \cite{UniLM}. For other languages, several variants of pre-trained models can be found such as AraBERT \cite{AraBERT} for Arabic, ChineseBERT \cite{ChineseBERT} for Chinese, DutchBERT \cite{DutchBERT} for Dutch, FrenchBERT \cite{FrenchBERT} for French, PhoBERT \cite{PhoBERT} for Vietnamese. However, there is one rising question for exploring multi-source transfer learning to borrow knowledge from multi-trained models. There have existed several approaches proposed to explore transfer learning under multi-sources. Zhang et al. \cite{MGNC} assumed to have $\vm$ word embeddings with corresponding dimensions and then join these at the final layer by simply concatenating to form the final feature vector. Yin and colleagues \cite{meta_embeddings} introduced an ensemble approach of combining different public embedding sets with the aim of learning meta-embeddings, utilizing a simple neural network or Singular Value Decomposition (SVD) to define a projection from the meta-embedding space to the known embeddings. 

\subsection{Multi-source Transfer Learning with Access to Source Data}\label{subsec:related-multi-source}
There have been different multi-source transfer learning methods used in natural language processing problems.
Chen et al. \cite{chen-etal-2019-multi-source} presented a mixture-of-experts (MoE) model \cite{MoE} to combine a set of language expert networks, one per source language, each responsible for learning language-specific features for that source language during training. 
Jian and colleagues \cite{Ni_2017} presented two weakly supervised directions for the cross-lingual named entity recognition (NER) with the assumption that there is no human annotation in a target language. By automatically creating labeled NER data for the target language using an annotation projection on selected corpora and projecting word embeddings from the target language to a source language, the proposed techniques bypassed three other weakly supervised approaches on the CoNLL data. 
Xingjian et al. \cite{Xingjian_2020} proposed a new deep transfer earning algorithm, namely XMixup, that could efficiently utilize the knowledge transfer from the source to the target domains for different classification tasks. The experimental results on six vision datasets showed the better performance and efficiency of the XMixup in comparison with several baseline algorithms.
Han and co-workers \cite{Guo_Pasunuru_Bansal_2020} investigated the multi-source domain adaptation for text classification using a new DistanceNet-Bandit model. The proposed method can utilize a multi-armed bandit controller that dynamically takes source domain data (labeled) among different source domains and combines the target domain data (unlabeled) to extract the feature representations during the training process and learn an optimal transfer from sources domains to the target domain.
Zheng et al. \cite{ijcai2017-311} studied the cross-domain sentiment classification using two parameter-shared adversarial memory networks that could utilize a set of labeled data and unlabeled data in a source domain to predict the polarity of unlabeled samples from the target domains. The proposed memory networks can automatically capture the associated important sentiment words using the attention mechanism without manual selection and share them in both source and target domains to minimize the classification error. The experiments showed that the proposed technique could outperform other state-of-the-art approaches on the Amazon reviews benchmark dataset. 
Stephen and colleagues \cite{mayhew-etal-2017-cheap} proposed an efficient approach for cross-lingual named entity recognition that could use a lexicon to translate annotated data available from different high-resource languages to a low-resource language. Using the newly translated data, it could learn the corresponding NER model in the target language. The method also outperformed other state-of-the-art NER results in seven languages.
Phillip et al. \cite{keung-etal-2019-adversarial} proposed a new adversarial learning scheme with multilingual BERTs for zero-resource cross-lingual text classification and named entity recognition. The method used English text (labeled) and unlabeled non-English text (unlabeled) during training and selected hyperparameters using English evaluation sets. The experimental results demonstrated the improved performance on the multilingual ML- Doc text classification and CoNLL 2002/2003 named entity recognition tasks. 

\subsection{Transfer Learning for Vietnamese Language}
For Vietnamese, there was little work about utilizing transfer learning for the Vietnamese sentiment analysis. Nguyen et al. \cite{finetune_bert_vietnamese} finetuned BERT models on Vietnamese datasets and showed experimental results those using BERT could slightly outperform other models using Glove and FastText. PhoBERT \cite{PhoBERT} (regarded as the first public large-scale monolingual language and the state-of-the-art model for the Vietnamese language) is trained on a very large corpus of Vietnamese and improves the state-of-the-art in multiple Vietnamese-specific NLP tasks. However, to the best of our knowledge, the problem of multi-source transfer learning for the Vietnamese sentiment classification has not been explored. Moreover, existing methods presented in Section~\ref{subsec:related-multi-source} assume access to the source domains' data, which may not be suitable in real-word scenarios due to privacy issues. Therefore, our work focuses on a more general setting where only pre-trained models on the source domains are available. Such a scenario has only been explored in vision applications~\cite{NEURIPS2019_6048ff4e} and not yet studied in the NLP domain.

\section{Methodology}
\label{sec:method}
In this section, we first formulate the sentiment classification problem and then describe the proposed \textbf{LIFA}, a simple yet effective method for integrating several pre-trained models to improve the performance on the sentiment classification task. LIFA makes a prediction of an input based on a novel gating mechanism to combine the embedding features from several experts, each of which may have {\bfseries different embedding sizes}. Moreover, LIFA allows for an easy mechanism to enforce certain structures to the experts, which results in three variants: (i) LIFA-COOP: experts cooperate with one another, (ii) LIFA-WTA: experts compete with one another, (iii) LIFA-SIGMOID: no specific structure.  


\subsection{Preliminary}
We first introduce the sentiment classification problem studied in this work. Let $\mc D = \{\vx_i, y_i \}_{i=1}^n$ be a training set consisting of $\{ \vx, y\}$, where $\vx = \{ x_1, \ldots, x_T \}$ is a training sequence $\vx$ of $T$ tokens $x_j$ and its corresponding label $y \in \{0,1\}$, which represents \emph{positive} (1) or \emph{negative} sentiment (0).

A classification model is composed of an embedding model $g(\cdot; \varphi)$ parameterized by $\varphi$, and a classifier $f(\cdot; \theta)$ parameterized by $\theta$. For simplicity, we omit the dependency of the embedding and classification models on their parameters $\varphi$ and $\theta$ in the rest of this paper. 
Given an input sequence $\vx$, a sequence of token-embeddings is first generated as: $\ve(\vx) = \{g(\vx_1),\ldots, g(\vx_T\}$. Then, the classifier takes the embedding tokens as input and makes a prediction as:
\begin{equation}
  \hat{y} = f \circ g (\vx) = f(\{g(\vx_1),\ldots, g(\vx_T \})
  \label{eqn:pred}
\end{equation}
Both the embedding parameters $\varphi$ and the classifier's parameters $\theta$ are jointly optimized by minimizing the empirical loss $\mc L(\hat{y},y)$, which is usually implemented as the cross-entropy loss for classification problems. In practice, the classifier $f$ is implemented as a deep neural network such as Recurrent Neural Networks and its variants~\cite{DRNN}. Meanwhile, the embedding model $g$ can be a pre-trained word embedding such as Word2vec \cite{Word2Vec}, GLOVE \cite{Glove}), or even complex pre-trained models (e.g., BERT \cite{BERT} or GPT \cite{GPT2}).

\subsection{LIFA: Multisource Transfer Learning for Vietnamese Sentiment Classification}
\begin{figure}
    \centering
    \includegraphics[width=0.81\textwidth]{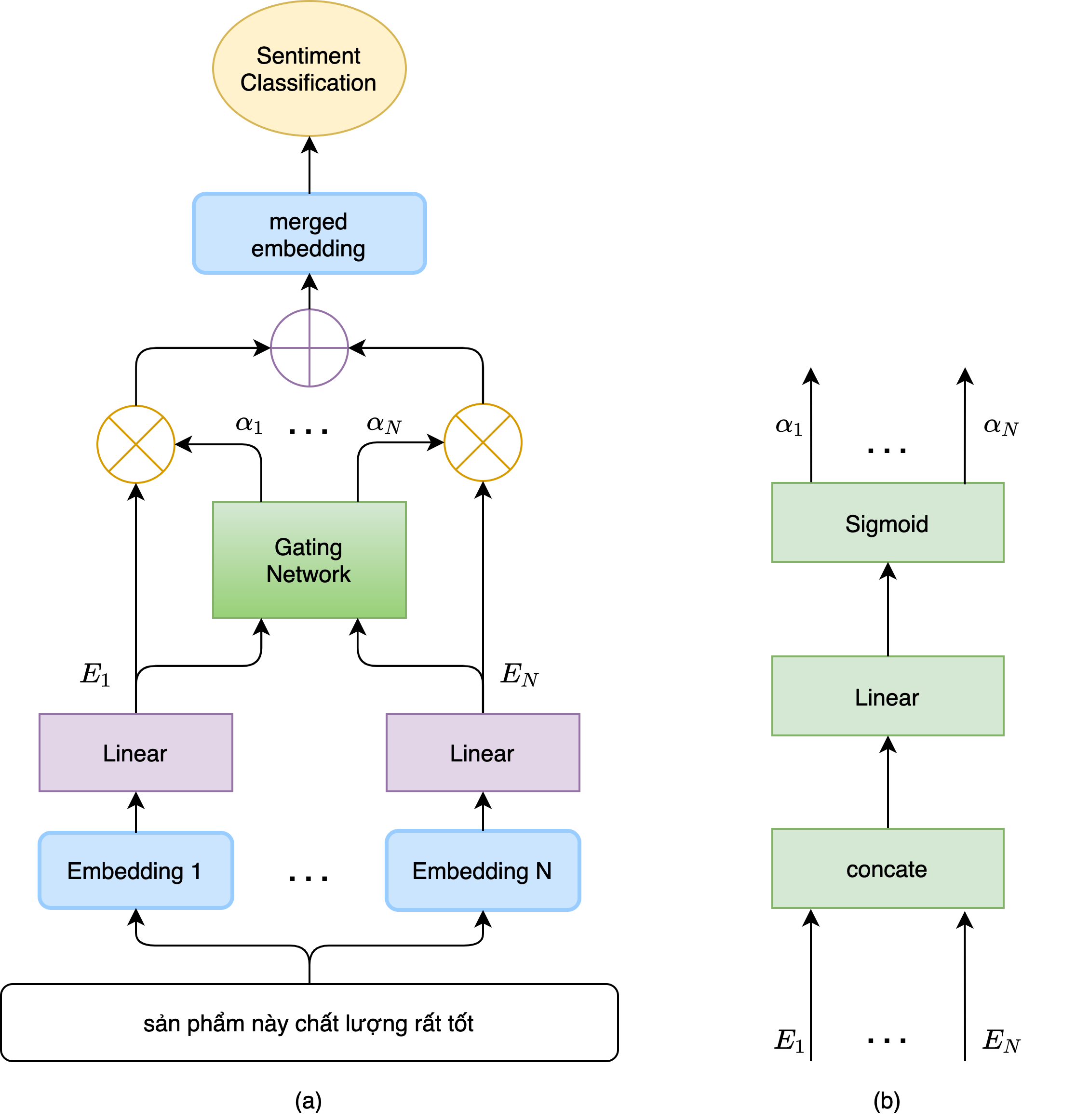}
    \caption{Our proposed LIFA framework using a gating network. (a) The overall framework demonstrated on $N$ embeddings. (b) Our Gating Network architecture. It is worth noting that from given input data, LIFA can select $N$ different embedding models (which can have distinct embedding dimensions) for extracting feature vectors. These feature vectors then go through linear transformation layers to project these feature vectors into the same feature space. Finally, a gating network can be employed for combining these newly computed features to learn an optimal classifier for the sentiment classification problem.}
    \label{fig:arch}
\end{figure}
One particular challenge often faced in practice is that it is usually costly to label and obtain adequate data to achieve satisfactory results. As a result, one can leverage the language structure through transfer learning schemes from pre-trained models to the embedding model $\varphi$. Such approaches are ubiquitous in practice and have been shown to improve performance significantly, especially when training data are limited. It gives rise to a multi-source transfer learning problem: given a set of pre-trained models (sources), how can one choose the appropriate model for transfer learning given a task at hand? 

This paper proposes LIFA (\textbf{L}earn\textbf{I}ng \textbf{F}or \textbf{A}lmagamation): a framework for tackling this multi-source transfer learning with pre-trained models problem. LIFA employs a Mixture-of-Experts (MoE) layer that learns to combine various sources by taking advantage of the knowledge from all of them. Therefore, LIFA can assign appropriate importance to each expert (pre-trained source model) such that the performance on the target problem is maximized. 

Moreover, we propose three variants of LIFA that enforce specific structures on the source knowledge: 
\begin{enumerate}[(i)]
\item LIFA-SIGMOID  (Learning for Amalgamation by Cooperation without any constraint) leads the experts working altogether without any constraint by nominating high weights to the experts having good performance and low weights to the experts having a poor performance.
\item LIFA-COOP (Learning for Amalgamation by Cooperation) forces the experts to cooperate with each other by smoothing out the weights among experts.
\item LIFA-WTA (Learning for Amalgamation by Winner-Take-All) encourages the competition among experts by assigning most of the weight to the expert having the best performance and almost zero weight to other experts.
\end{enumerate}

\subsubsection{LIFA-SIGMOID}
We assume having access to $n$ embedding models $\bm E_1, \ldots, \bm E_n$ with different embeddings' dimensions. Our LIFA first applies linear transformation layers on the embedding models $\{\bm E_1, \bm E_2, ..., \bm E_n\}$ to obtain new embedding models $\{\bm E_1^{'}, \bm E_2^{'}, ..., \bm E_n^{'}\}$ that have same dimensionality of K, which can be formulated as follows:
\begin{equation}
	\bm E_i^{'}(\vx) = \bm W_i\bm E_i(\vx),
  \label{eqn:new_embs}
\end{equation}
where $\bm W_i$ is the parameter matrix of the i-th linear layer. We then compute the final embedding as a weighted combination of the experts' embeddings: 
\begin{equation}
  \bm E(\vx) = \sum_{i=1}^n \bm \alpha_{i} \bm E_i^{'}(\vx).
  \label{eqn:output}
\end{equation}
Here, $G(\vx)$ is the gating network that can learn to combine the embeddings to predict importance coefficients $[\bm \alpha_{0}, \bm \alpha_{1}, ..., \bm \alpha_{n}]$ of each expert. In our experiments, we have up to three experts for each classification problem of Vietnamese or English. In this work, the gating network receives the embeddings $\{\bm E_1^{'}, \bm E_2^{'}, ..., \bm E_n^{'}\}$ as the input data and generates an output as a sparse n-dimensional vector. We implement the gating network by multiplying the concatenation of these embeddings by a trainable weight matrix $W$ and then apply the $Sigmoid$ function, which can be formulated as follows:
\begin{equation}
  G(\vx) = Sigmoid([\bm E_1^{'}(\vx), \bm E_2^{'}(\vx), ..., \bm E_n^{'}(\vx)]\bm W).
  \label{eqn:gating}
\end{equation}
Since there is no constraints on the experts' weights, each expert's weight will be updated proportionally to its contribution in the final prediction. Therefore, this variance of LIFA is named LIFA-SIGMOID. Once acquiring the consolidated embedding vector $\bm E(\vx)$, we can make a prediction $\hat{y}$ in the same manner as Eq. (\ref{eqn:pred}). Regarding the sentiment classification module $\bm \theta$, we use a linear layer with the output features of ``2'', indicating the number of sentiment polarities. Figure~\ref{fig:arch} demonstrates the workflow of our LIFA and the architecture of our gating layer.

\subsubsection{LIFA-COOP}
We now describe a simple and effective technique to enforce a prior structure to expert. Particularly, we are interested in \emph{two} specific structures in which the experts cooperate or compete with one another. Thanks to the gating network's design, we can easily achieve this goal by changing the gating activation function from $Sigmoid$ to $Softmax$ in Eq.~\ref{eqn:gating} as follows:
\begin{equation}\label{lifa-softmax}
  G(x) = \textrm{Softmax}_{\tau}([\bm E_1^{'}(\vx), \bm E_2^{'}(\vx), ..., \bm E_n^{'}(\vx)]\bm W),
\end{equation}
where $\textrm{Softmax}_{\tau}: \mathbb{R}^K \rightarrow \mathbb{R}^K$ is the standard $Softmax$ function with temperature $\tau$ defined over $K$ inputs $z_1,z_2,\ldots,z_K$ as:
\begin{equation}\label{softmax}
  \textrm{Softmax}_{\tau}(z_i) = \frac{\exp ^{z_i/\tau}}{\sum_{j=1}^K \exp ^{z_j/\tau}}.   
\end{equation}
It is important to note that as $\tau \rightarrow \infty$, we have $\textrm{Softmax}_{\tau}(z_i) \approx \textrm{Softmax}_{\tau}(z_j), \forall i,j$, i.e., the $Softmax$ function with high temperatures will produce a uniform distribution over its input. Conversely, when $\tau \rightarrow 0$, the $Softmax$ function will converge to a Dirac delta distribution peaking at the largest value input, i.e. $\arg\max_i{z_i}$. Therefore, we propose implementing the LIFA variants by first replacing the $Sigmoid$ activation in LIFA's gating network with the $Softmax$ function. For LIFA-COOP, we raise the temperature $\tau$ to a high value so that experts will receive similar weight signals regardless of their performance. 

\subsubsection{LIFA-WTA}
LIFA-WTA has the same architecture as LIFA-COOP. However, we instead lower the temperature $\tau$ so that only the experts making the correct prediction receive most of the rewards, reflecting the winner-take-all principle. In what follows, we also compare the performance of different variants of LIFA in chosen datasets. 

\section{Experiments}
\label{sec:exp}
We design our experiments to investigate the following hypotheses: (i) it is more beneficial to take transfer knowledge from several pre-trained models compared to using just one (Section~\ref{subsec:viet} and~\ref{subsec:eng}); (ii) our LIFA framework can efficiently transfer knowledge compared to the naive strategy of concatenating all the embeddings (Sections~\ref{subsec:viet} and~\ref{subsec:eng}); and (iii) LIFA allows for a flexible mechanism to control to which degree pre-trained knowledge should be utilized (Sections~\ref{subsec:gating_size}) and a mechanism to enforce prior structure to the experts (Section~\ref{subsec:temp}). 

\subsection{Datasets}
\begin{table}[t]
\centering
\begin{tabular}{|c|c|c|| c|}
\hline
\textbf{AIVIVN Dataset} & Positive & Negative & Total \\ \hline
Train & 8690 & 7383 & 16073\\ \hline
Test & 5767 & 5214 & 10981\\ \hline
\hline
\textbf{AISIA-VN-Review-S  Dataset} & Positive & Negative & Total \\ \hline
Train 5K & 3912 & 1088 & 5000\\ \hline
Train 15K & 11736 & 3264 & 15000\\ \hline
Train 25K & 19559 & 5441 & 25000\\ \hline
Test & 137833 & 30210 & 168043\\ \hline
\end{tabular}
\caption{Two Vietnamese datasets used in our experiments.}
\label{table:dataset_vietnamese}
\end{table}

\begin{table}[t]
\centering
\begin{tabular}{|c|c|c|| c|}
\hline
\textbf{Multi-Domain Dataset} & Positive & Negative & Total \\ \hline
Books & 1000 & 1000 & 2000\\ \hline
DVD & 1000 & 1000 & 2000\\ \hline
Electronics & 1000 & 1000 & 2000\\ \hline
Kitchen and Housewaves & 1000 & 1000 & 2000\\ \hline
\hline
\textbf{Amazon Review Dataset} & Positive & Negative & Total \\ \hline
Cell Phones and Accessories & 10000 & 10000 & 20000\\ \hline
Clothing Shoes and Jewelry & 10000 & 10000 & 20000\\ \hline
Home and Kitchen & 10000 & 10000 & 20000\\ \hline
Tools and Home Improvement & 10000 & 10000 & 20000\\ \hline

\end{tabular}
\caption{Two English datasets used in our experiments.}
\label{table:dataset_english}
\end{table}
We consider four different datasets in our experiments. Two Vietnamese datasets are used for the Vietnamese sentiment classification, including the \textbf{AIVIVN dataset} and the \textbf{AISIA-VN-Review-S} (a subset of the AISIA-Review-F). We also extend to English sentiment analysis on two multi-domains datasets: the \textbf{Multi-Domain Dataset}~\cite{MultiDomainDataset} and the \textbf{Amazon Review Dataset}~\cite{AmazonReviewDataset}. The statistics of each dataset are shown in Tables \ref{table:dataset_vietnamese} and \ref{table:dataset_english}. In the following, we provide the details of the aforementioned datasets.

\textbf{AIVIVN Dataset} is a Vietnamese review dataset that consists of around 16K reviews in the training set and around 11K reviews in the testing set. This dataset was used for the Vietnam Sentiment Analysis Challenge 2019\footnote{https://www.aivivn.com/contests/1}. All labels in the testing set are not available and kept private from the competition organizers. We carefully labeled all the testing reviews by ourselves and cross-checked multiple times between our team members and experts to guarantee the labeling quality. Along with this work, we will publish this dataset with our label for further research. 

In \textbf{AISIA-VN-Review-S} and \textbf{AISIA-VN-Review-F} datasets, we first collect 450K customer reviewing comments from various e--commerce websites. Then, we manually label each review to be either positive or negative, resulting in 358,743 positive reviews and 100,699 negative reviews. We named this dataset the sentiment classification from reviews collected by AISIA, the full version (AISIA-VN-Review-F). However, in this work, we are interested in improving the model's performance when the training data are limited; thus, we only consider a subset of up to 25K training reviews and evaluate the model on another 170K reviews. We refer to this subset from the full dataset as AISIA-VN-Review-S. 
It is important to emphasize that our team spends a lot of time and effort to manually classify each review into positive or negative sentiment. To the best of our knowledge, AISIA-VN-Review-F is the most extensive large-scale dataset for Vietnamese sentiment analysis until now, which can be considered as an additional contribution to the research community in Vietnamese Natural Language Processing. Due to our text data collected from social networks and e-commerce websites, we observe that there exist many informal texts and words that do not conform to the usual standard of the Vietnamese language. Thus, we apply various pre-processing steps for the text data, as described in Table \ref{table:pre_processing}. 
Lastly, although some of our pre-processing steps such as Step \#2 may potentially remove additional information about emotions, we choose to standardize all the writing styles in this work for a consistent comparison across all datasets. We will also publish the unprocessed dataset to facilitate future research exploring such properties.


\begin{table}
\centering
\scalebox{0.96}{
\begin{tabular}{| P{0.05\linewidth} | p{0.2\linewidth} | p{0.5\linewidth} | p{0.25\linewidth}|}
\hline
Step & \multicolumn{1}{|c|}{Description} & \multicolumn{1}{|c|}{Customer Review (Raw)} & \multicolumn{1}{|c|}{Post-processed Text}\\ \hline
1 & Lowercase all characters & mua xong về bỏ sọt rác luôn. màn hình kính thì mờ, mặt kính thì trầy xước, nhìn không khác gì đồ cũ phế liệu đem bán cho khách hàng. \textbf{NẾU MÀ TIẾP TỤC XEM CÁI NÀY CHẮC LÀ HƯ LUÔN CON MẮT}... \textbf{BỰC MÌNH}... \hfill \break \textit{we throw it into the trash after buying due to the glass screen is blur and scratched, so it looks like a garbage which sold to customer. If continuously watching this screen, it will harm our eyes ... so angry}  & mua xong về bỏ sọt rác luôn. màn hình kính thì mờ, mặt kính thì trầy xước, nhìn không khác gì đồ cũ phế liệu đem bán cho khách hàng. \textbf{Nếu mà tiếp tục xem cái này chắc là hư luôn con mắt}... \textbf{bực mình}...\\ \hline

2 & Correct elongated words & Giao hàng nhanh hơn dự kiến, vải \textbf{đẹpppppppppppppppppp}! \hfill \break \textit{The delivery is faster than expected, the fabric is so beautifulllllllllll!} & giao hàng nhanh hơn dự kiến vải \textbf{đẹp}\\ \hline

3 & Remove URLs & Các bác tham khảo ở đây, rẻ hơn hẳn 100-150k \textbf{https://noithatluongson.vn/ban-chan-sat} \hfill \break \textit{you guys can refer here, cheaper than 100-150k} & Các bác tham khảo ở đây, rẻ hơn hẳn 100-150k\\ \hline

4 & Translate & Mình đặt chiều hôm qua đến sáng nay thì có hàng rồi. Nhanh hú hồn. \textbf{Thanks} \textbf{shop} nhé \hfill \break \textit{I ordered yesterday but we received the product today. So fast.} & mình đặt chiều hôm qua đến sáng nay thì có hàng rồi nhanh hú hồn \textbf{cảm ơn} \textbf{cửa hàng} nhé\\ \hline

5 & Remove punctuation marks and special characters & Hàng đúng chuẩn\textbf{,} đóng gói cẩn thận\textbf{,} dùng tốt \textbf{,} ủng hộ\textbf{!} \hfill \break \textit{The product is nice, the packaging is careful, the usage is good.} & Hàng đúng chuẩn đóng gói cẩn thận dùng tốt ủng hộ\\ \hline

6 & Exclude other language reviews (Korean, Chinese, English, etc.) & The quality is good and suitable for using at the library, but the click is not good. & -\\ \hline


7 & Correct freestyle letters and acronyms & dày , êm , rất tốt so với giá tiền . hình in trên miếng lót rất chi tiết và rõ nét . xài \textbf{tgian} thì sẽ đánh giá thêm \hfill \break \textit{the product is thick, smooth, deserved with its price. It is very detailed and clear. I will feedback more after usage} & dày , êm , rất tốt so với giá tiền. hình in trên miếng lót rất chi tiết và rõ nét. xài \textbf{thời gian} thì sẽ đánh giá thêm\\ \hline 

\end{tabular}}
\caption{An illustration for the preprocessing step in \textbf{AISIA-VN-Review-F  Dataset}.}
\label{table:pre_processing}
\end{table}

\textbf{Multi-Domain Dataset} \cite{MultiDomainDataset} consists of a short English dataset from four different domains of books, DVD, electronics, kitchen, and housewares, taken from Amazon.com. Also, each domain contains 1K positive reviews and 1K negative reviews. 

\textbf{Amazon Review Dataset} \cite{AmazonReviewDataset} is a very large-scale English review dataset with 19 different categories and millions of reviews. In this paper, we follow \cite{gatedCNN} and select a subset of four domains and reviews: Cell Phones and Accessories, Clothing Shoes and Jewelry, Home and Kitchen, Tools and Home Improvement. In each domain, we randomly select a subset of 20K reviews divided equally among the positive and negative sentiment.

\subsection{Baselines}
We compare our LIFA with various baselines, from individual models to classical ensemble methods as described below.

\textbf{Recurrent CNN} \cite{RCNN} proposes a combination of Recurrent and Convolutional Neural Network for text classification and shows its remarkable improvement compared to the individual RNNs or CNNs. The model takes advantage of RNN to capture long-term dependencies and contextual information and CNN to extract local and position-invariant features very well. We implement this model from scratch with the same settings as presented in the original paper.


\textbf{BERT, PhoBERT, XLM} The second set of baselines we consider is transformer-based models. Particularly we consider the pre-trained models of $BERT_{base-multilingual}$ \cite{BERT}, $PhoBERT_{base}$ \cite{PhoBERT},
and XLM \cite{XLM} for our task. $BERT_{base-multilingual}$ is pre-trained on cased text on the top 104 languages with the largest Wikipedias, while $PhoBERT_{base}$ is the first public large-scale monolingual language model pre-trained for Vietnamese. These pre-trained models are also regarded as experts for our proposed LIFA.

\textbf{Concatenation} A traditional method of ensembling multiple pre-trained expert models~\cite{MGNC} in which the experts' embeddings are concatenated before feeding into the fully-connected layers.


\subsection{Implementation}
All experiments are performed on a deep learning workstation with Intel Core i9-7900X CPU, 128GB RAM and two GPUs RTX-2080Ti with Pytorch framework \cite{paszke2019pytorch}.
For Recurrent CNN~\cite{RCNN}, we train the model from scratch with the feature embedding of dimension 256 and the Fasttext word embedding vectors\footnote{https://fasttext.cc/docs/en/crawl-vectors.html} with dimension of 300.
For $PhoBERT_{base}$, we first apply the Vietnamese word segmenter RDRsegmenter \cite{ViWordSegmentor} to process raw data and generate segmented words, then we employ a pre-trained model\footnote{https://github.com/VinAIResearch/PhoBERT} and fine-tune on all datasets. The dimension of feature embedding from the PhoBERT model is 768. For $BERT_{base-multilingual}$ and XLM, we utilizes the pre-trained models\footnote{https://huggingface.co/transformers/} and fine-tune on all datasets. The dimension of feature embedding from BERT and XLM are 768 and 1024, respectively. For LIFA, we consider three component experts: Recurrent CNN, BERT, and PhoBERT for the Vietnamese datasets; Recurrent CNN, BERT, and XLM for the English datasets. We train the component experts individually and store the best checkpoint, which is then used as the experts to train LIFA. All methods are optimized to minimize the cross-entropy loss using the Adam optimizer with batch size of eight over 30 epochs with early stopping of five based on the validation accuracy. Lastly, we compared the methods over three evaluation metrics: AUC, Accuracy, and F1-score.

\subsection{Experimental results on Vietnamese review datasets}\label{subsec:viet}
\label{sec:exp_Vietnamese}
\paragraph{Standard evaluation} Tables \ref{table:result_aivivn_dataset} and \ref{table:result_our_dataset} report the experimental results on the AIVIVN and AISIA-VN-Review-S Datasets. It is worth noting that all methods considered in this work perform much better than the AIVIN 2019's champion with respect to our label test set. For our baselines, we observe that BERT and PhoBERT performs slighly better than Recurrent CNN, thanks to their stronger transformer backbone. Moreover, the Concatenation baseline performs slightly better than the single model, suggesting that taking advantage of multiple pre-trained models can potentially improve the performance. However, the Concatenation strategy is quite simple and may not efficiently utilize the rich knowledge of all experts, making it performs worse than our LIFA strategies.
Overall, we observe that two LIFA variants (LIFA-COOP and LIFA-SIGMOID) consistently outperform the remaining methods on both datasets across all metrics.
For LIFA, we observe that LIFA-SIGMOID achieves the best performance, while LIFA-WTA performs the worst. This result suggests that LIFA-SIGMOID achieves a great flexibility in knowledge sharing experts while enforcing prior structure to the experts does not perform well in all scenarios.


\paragraph{Performance with increasing number of training data} When the number of training samples in the target domain is limited, it is essential to rely on the knowledge stored in the pre-trained models. Therefore, we also report the results of our AISIA-VN-Review-S Dataset in Table~\ref{table:result_our_dataset} with three different amounts of training data: 5K, 15K, and 25K, while keeping the same testing dataset of 170K reviews. 
Generally, the performance improves with more training data across all methods, which is easy to understand as deep models require a large amount of training data to achieve good performance. Interestingly, the performance gap between our LIFA-SIGMOID compared to other methods is most significant with 5K reviews and becomes stable when moving to 15k and 25k reviews. This result shows that our LIFA-SIGMOID can efficiently utilize pre-trained knowledge under limited training samples while still maintains its ability to adapt with more training data.

\begin{table}[ht]
\centering
\begin{tabular}{|l|c|c|c|}
\hline
\textbf{Methods} & AUC & ACC & F1 \\ \hline
AIVIVN 2019 Sentiment Champion & - & - & 90.01 \\ \hline
Recurrent CNN & 98.33 & 93.42 & 92.98 \\ \hline
BERT & 98.82 & 94.05 & 93.94 \\ \hline
PhoBERT & 98.67 & 94.04 & 93.79 \\ \hline
Concatenation & 98.12 & 94.37 & 94.09 \\ \hline
LIFA-WTA & 98.04 & 93.41 & 93.02 \\ \hline
LIFA-COOP & 99.02 & 95.11 & 94.87 \\ \hline
LIFA-SIGMOID & \textbf{99.12} & \textbf{95.46} & \textbf{95.20} \\ \hline
\end{tabular}
\caption{The results of our proposed LIFA and other methods (including the AIVIVN 2019 Sentiment Champion's solution) on \textbf{AIVIVN Dataset}. Here, we consider three different performance metrics: Accuracy (ACC), AUC, and F1-score (F1).}
\label{table:result_aivivn_dataset}
\end{table}

\begin{table}[t]
  \centering
  \renewcommand{\arraystretch}{1.2}
  \scalebox{0.96}{
 \begin{tabular}{|c|c|c|c|c|c|c|c|c|c|}
    \hline
   \multirow{2}{*}{\textbf{Methods}} & \multicolumn{3}{c|}{\textbf{5K reviews}} & \multicolumn{3}{c|}{\textbf{15K reviews}} & \multicolumn{3}{c|}{\textbf{25K reviews}} \\
    \cline{2-10}
    & AUC & ACC & F1 & AUC & ACC & F1 & AUC & ACC & F1\\
    \hline
    Recurrent CNN & 90.10 & 86.57 & 66.53 & 93.35 & 88.92 & 74.35 & 93.83 & 89.57 & 75.00\\ \hline
    BERT & 90.01 & 87.41 & 67.43 & 94.09 & 89.45 & 76.03 & 94.62 & 90.00 & 75.80\\ \hline
    PhoBert & 91.95 & 88.04 & 68.98 & 94.07 & 89.66 & 75.25 & 94.62 & 90.57 & 77.65\\ \hline
    Concatenation & 92.57 & 88.14 & 69.26 & 94.49 & 90.24 & 77.89 & 95.17 & 90.6 & 77.57\\ \hline
    LIFA-WTA & 92.05 & 88.06 & 69.00 & 94.42 & 90.04 & 76.82 & 94.57 & 90.81 & 78.30\\ \hline
   LIFA-COOP & 92.70 & 88.11 & 69.36 & 95.39 & 91.18 & 79.18 & 95.16 & 91.07 & 79.21\\ \hline
   LIFA-SIGMOID & \textbf{92.95} & \textbf{88.58} & \textbf{69.55} & \textbf{95.71} & \textbf{91.42} & \textbf{79.83} & \textbf{95.79} & \textbf{91.76} & \textbf{80.18}\\ \hline
  \end{tabular}}
  \caption{The results of our proposed LIFA and other methods on \textbf{AISIA-VN-Review-S  Dataset} with 10K, 15K and 20K training reviews consecutively. Here, we consider three different performance metrics: Accuracy (ACC), AUC, and F1-score (F1).}
  \label{table:result_our_dataset}
\end{table}

\begin{figure}[t]
    \centering
    \includegraphics[width=1.0\textwidth]{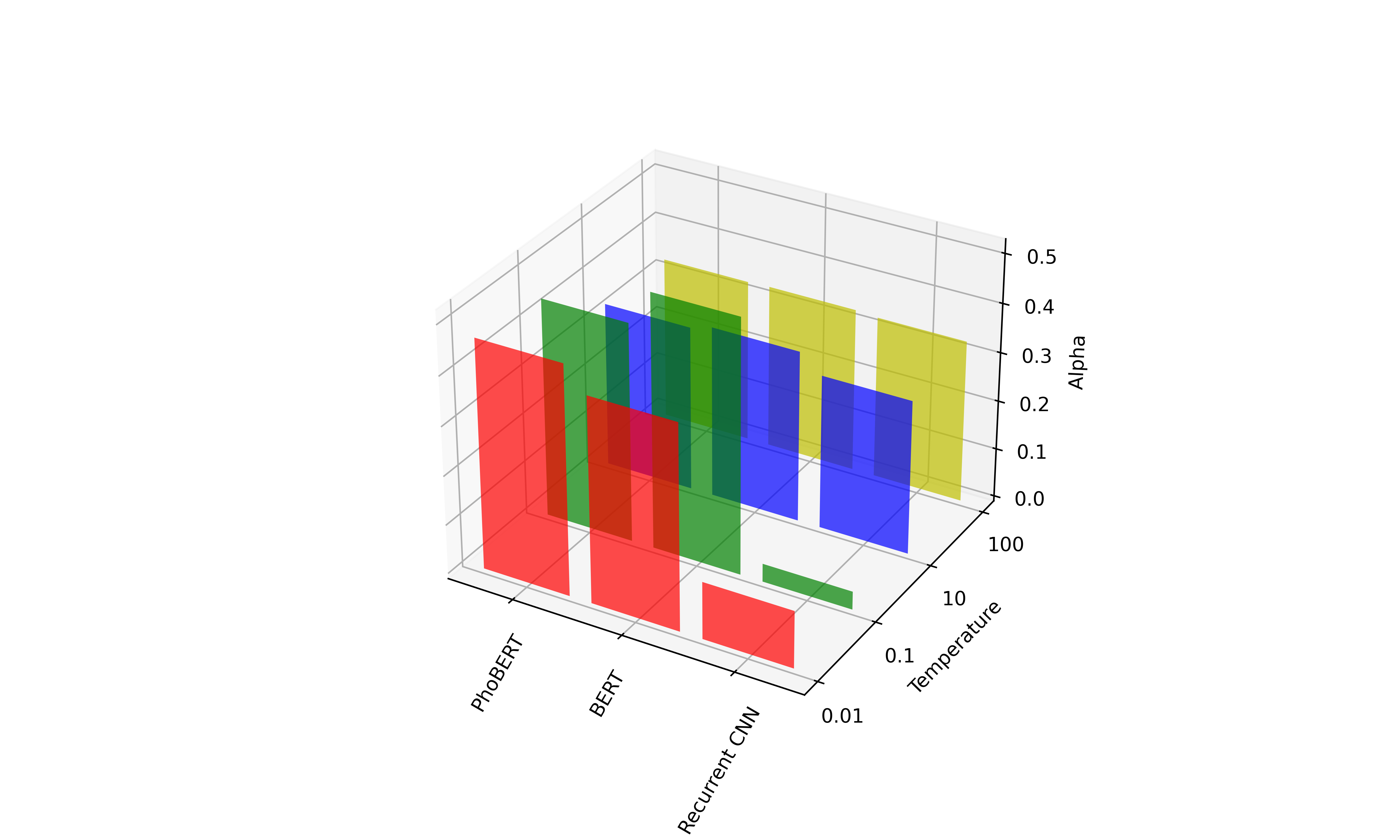}
    \caption{The weight distribution learnt by our LIFA with different Softmax temperatures of 0.01, 0.1, 10 and 100 on \textbf{AIVIVN} Dataset.}
    \label{fig:weight_distribution_aivivn}
\end{figure}
\begin{figure}[t]
    \centering
    \includegraphics[width=1.0\textwidth]{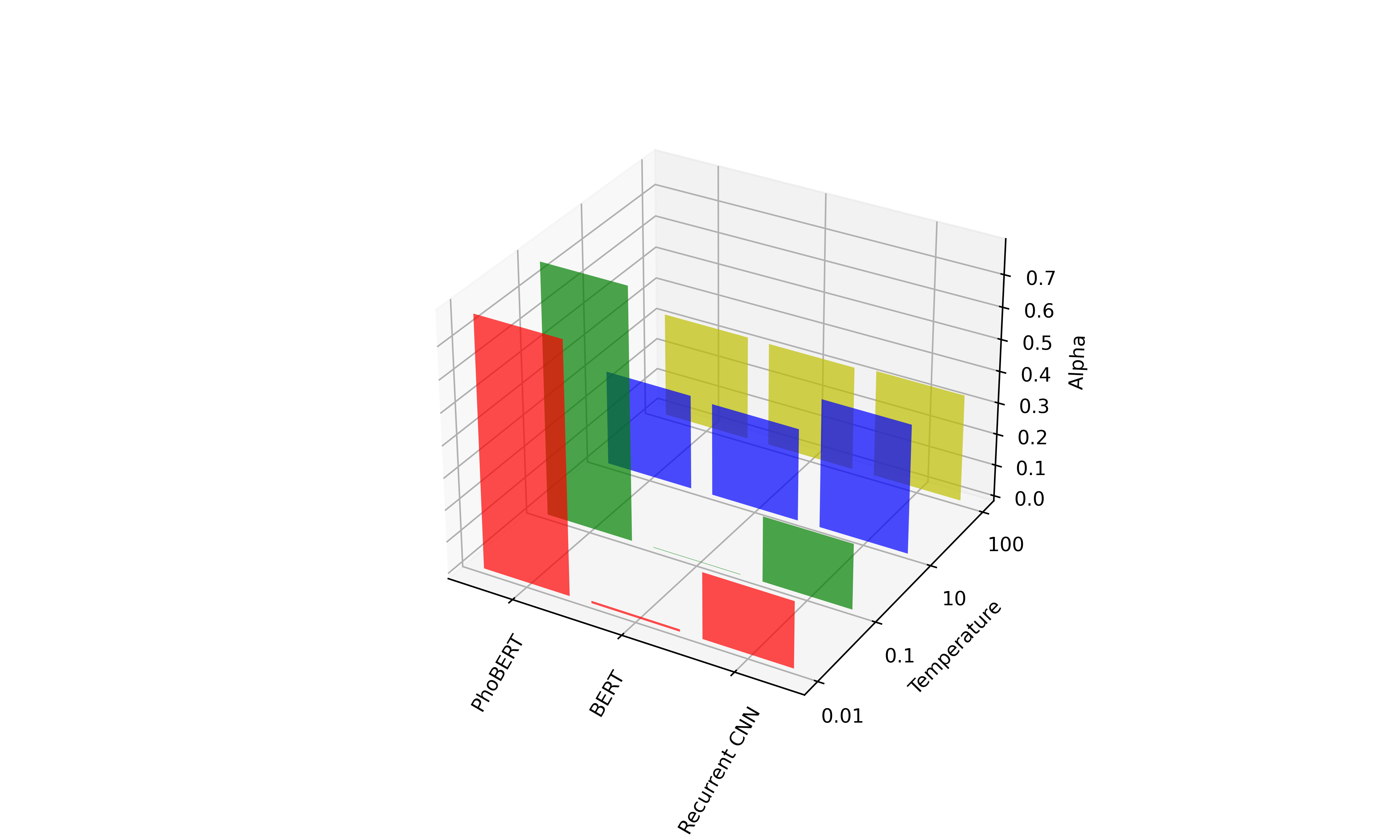}
    \caption{The weight distribution learnt by our LIFA with different Softmax temperatures of 0.01, 0.1, 10 and 100 on \textbf{AISIA-VN-Review-S} Dataset.}
    \label{fig:weight_distribution_tiki}
\end{figure}

\subsection{Experimental results on English review datasets}\label{subsec:eng}
\label{sec:exp_English}
Similarly, we conduct the experiments of the two English multi-domains datasets (``Multi-Domain Dataset'' and ``Amazon Reviews Dataset'') \cite{gatedCNN}. We employ LIFA-SIGMOID, the best one in LIFA variants, and compare with other baselines, as mentioned in \cite{gatedCNN}. It is worth noting that GLU, GTU, and GTRU are three state-of-the-art models for those two datasets \cite{gatedCNN}. The experiments on these multi-domain datasets require training the models on one domain and testing them on other remaining domains. Tables \ref{table:result_multi_domain_dataset} and \ref{table:result_amazon_reviews_dataset} present the experimental results of this experiment.

Consistent with the previous experiments on the Vietnamese datasets. Here we also observe that for the single expert methods such as GLU, GTU, BERT, etc., the performance increases with better backbones: BERT and XLM outperform the other baselines. Second, Concatenation and LIFA-SIGMOID consistently performs better than the remaining baselines thanks to the knowledge from multiple experts. Lastly, our LIFA-SIGMOID achieves the best results, outperforms the Concatenation, by efficiently sharing knowledge across experts.

\begin{sidewaystable}
\renewcommand{\arraystretch}{1.5}
  \scalebox{0.9}{
\begin{tabular}{|l|c|c|c|c|c|c|c|c|}
\hline
\textbf{Source $\rightarrow$ Target} & Recurrent CNN & BERT & XLM & Concatenation & LIFA-SIGMOID & GLU \cite{gatedCNN} & GTU \cite{gatedCNN} & GTRU \cite{gatedCNN}\\ \hline
Books $\rightarrow$ DVD & 77.95 & 79.80 & 79.40 & 82.35 & \textbf{82.75} &79.50 & 79.25&79.25\\ \hline
Books $\rightarrow$ Electronics & 74.45 & 75.15 & 77.35 & 78.20 & \textbf{79.40} & 71.75& 71.75&71.75\\ \hline
Books $\rightarrow$ Kitchen & 76.35 & 79.70 & 79.85 & 82.80 & \textbf{83.55} & 73.00&72.50 &72.50\\ \hline
DVD $\rightarrow$ Books & 75.05 & 77.60 & 79.15 & 82.65 & \textbf{83.65} &78.00 & 80.25&77.25\\ \hline
DVD $\rightarrow$ Electronics & 73.60 & 76.65 & 78.80 & 79.70 & \textbf{81.60} & 73.00& 74.50&69.25\\ \hline
DVD $\rightarrow$ Kitchen & 74.20 & 78.90 & 79.75 & 82.00 & \textbf{83.85} &77.00 &76.00 &74.75\\ \hline
Electronics $\rightarrow$ Books & 70.15 & 77.05 & 70.15 & 81.35 & \textbf{81.35} & 71.75& 68.75&67.25\\ \hline
Electronics $\rightarrow$ DVD & 70.60 & 75.10 & 76.40 & 79.80 & \textbf{80.40} & 71.75& 69.00&68.25\\ \hline
Electronics $\rightarrow$ Kitchen & 80.90 & 83.90 & 85.50 & 88.60 & \textbf{89.10} & 82.25& 82.25&79.00\\ \hline
Kitchen $\rightarrow$ Books & 71.50 & 75.40 & 78.45 & 80.00 & \textbf{81.25} &70.00 & 67.75&63.25\\ \hline
Kitchen $\rightarrow$ DVD & 72.30 & 73.40 & 75.30 & 76.55 & \textbf{78.25} & 73.75& 73.50&69.25\\ \hline
Kitchen $\rightarrow$ Electronics & 78.10 & 80.85 & 82.90 & 84.85 & \textbf{85.55} & 82.00& 82.00&81.25\\ \hline
\end{tabular}}
\caption{The accuracy between our proposed LIFA and the baselines on the \textbf{Multi-Domain Dataset}.}
\label{table:result_multi_domain_dataset}
\end{sidewaystable}

\begin{sidewaystable}[ht]
\renewcommand{\arraystretch}{1.5}
  \scalebox{0.9}{
\begin{tabular}{|l|c|c|c|c|c|c|c|c|}
\hline
\textbf{Source $\rightarrow$ Target} & Recurrent CNN & BERT & XLM & Concatenation & LIFA-SIGMOID & GLU \cite{gatedCNN}& GTU \cite{gatedCNN}& GTRU \cite{gatedCNN}\\ \hline
Cell Phone $\rightarrow$ Clothing & 83.88 & 87.61 & 88.85 & 89.87 & \textbf{90.06} & 85.13&84.95 &84.80\\ \hline
Cell Phone $\rightarrow$ Home & 88.03 & 89.17 & 92.17 & 92.33 & \textbf{93.03} &84.85 &84.20 &84.55\\ \hline
Cell Phone $\rightarrow$ Tools & 88.09 & 90.14 & 91.75 & 92.62 & \textbf{93.13} & 79.50& 79.28&80.23\\ \hline
Clothing $\rightarrow$ Cell Phone & 84.69 & 87.06 & 87.94 & 88.93 & \textbf{89.04} &80.93 &80.25 &83.10\\ \hline
Clothing $\rightarrow$ Home & 89.70 & 90.45 & 91.61 & 92.16 & \textbf{92.25} &83.95 &83.40 &84.03\\ \hline
Clothing $\rightarrow$ Tools & 88.65 & 88.50 & 90.47 & 90.78 & \textbf{90.96 } &79.48 &77.85 &79.38\\ \hline
Home $\rightarrow$ Cell Phone & 86.89 & 86.89 & 90.79 & 90.72 & \textbf{90.92} &83.18 &81.85 &82.10\\ \hline
Home $\rightarrow$ Clothing & 85.87 & 89.43 & 91.64 & 91.94 & \textbf{92.14} &82.75 &84.10 &85.43\\ \hline
Home $\rightarrow$ Tools & 89.94 & 91.73 & 93.36 & 94.08 & \textbf{94.19} &82.55 &81.78 &81.83\\ \hline
Tools $\rightarrow$ Cell Phone & 84.84 & 88.24 & 89.36 & 89.97 & \textbf{90.28} &82.13 & 80.81&81.83\\ \hline
Tools $\rightarrow$ Clothing & 87.96 & 89.86 & 90.96 & 91.48 & \textbf{91.79} & 82.63&83.98 &84.78\\ \hline
Tools $\rightarrow$ Home & 87.54 & 91.33 & 91.4 & 92.52 & \textbf{92.94} &84.70 & 83.95&85.28\\ \hline
\end{tabular}}
\caption{The accuracy between our proposed LIFA and the baselines on \textbf{Amazon Reviews Dataset}.}
\label{table:result_amazon_reviews_dataset}
\end{sidewaystable}

\subsection{Ablation Study}
In this section, we conduct various ablation studies to demonstrate the robustness of LIFA under different embedding sizes and have a better understanding of the LIFA-COOP and LIFA-WTA behaviours.

\begin{table}[htbp]
  \centering
  \renewcommand{\arraystretch}{1.2}
 \begin{tabular}{|c|c|c|c|c|c|c|}
    \hline
   \multirow{2}{*}{\textbf{Methods}} & \multicolumn{3}{c|}{\textbf{AIVIVN Dataset}} & \multicolumn{3}{c|}{\textbf{AISIA-VN-Review-S  Dataset}} \\
    \cline{2-7}
    & AUC & ACC & F1 & AUC & ACC & F1\\
    \hline
    LIFA 256 & 98.95 & 95.12 & 94.83 & 94.99 & 90.54 & 78.12 \\ \hline
    LIFA 512 & \textbf{99.12} & \textbf{95.46} & \textbf{95.20} & \textbf{95.71} & \textbf{91.42} & \textbf{79.83} \\ \hline
    LIFA 768 & 98.85 & 94.98 & 94.68 & 94.9 & 90.47 & 77.71 \\ \hline
  \end{tabular}
  \caption{The experimental results of the proposed LIFA with different dimensions on the {\bf AIVIVN Dataset} and {\bf AISIA-VN-Review-S  Dataset}.}
  \label{table:study_dimension}
\end{table}

\subsubsection{Different Gating embedding sizes}\label{subsec:gating_size}
Individual experts play a vital role in our LIFA framework. However, different pre-trained models can provide different embedding dimensions, i.e. 256, 768, and 768 in our Vietnamese datasets. Therefore, LIFA employs a linear layer to map such embeddings to the same dimensions before they can be combine together. In this section, we investigate the effect of the common mapping dimension to the final performance. We consider the ``AIVIVN'' and ``AISIA-VN-Review-S`` datasets with the LIFA-SIGMOID model and vary the common mapping sizes, from 256, 512, to 768. The experimental results in Table~\ref{table:study_dimension} show that the increase of this dimension {\bf does not consistently improve} the performance of LIFA-SIGMOID. On both datasets, the performances increases when the common mapping size increases from 256 to 512, but then decreases when we further increase the mapping size to 768. One possible explaination is that the common mapping size is the bottleneck to transfer the knowledge from the pre-trained models to the current task. Small mapping size (256) limites the knowledge transfer to learn the current task. On the other hand, larger mapping size allows for more knowledge, but not all of them are useful, especially with limited training data. As a result, controlling the common mapping size in LIFA enables a flexible knowledge transfer mechanism to facilitate training across different datasets with different amount of training data.

\subsubsection{Cooperative and Competitive LIFA}\label{subsec:temp}
\begin{figure}[t]
    \centering
    \includegraphics[width=0.99\textwidth]{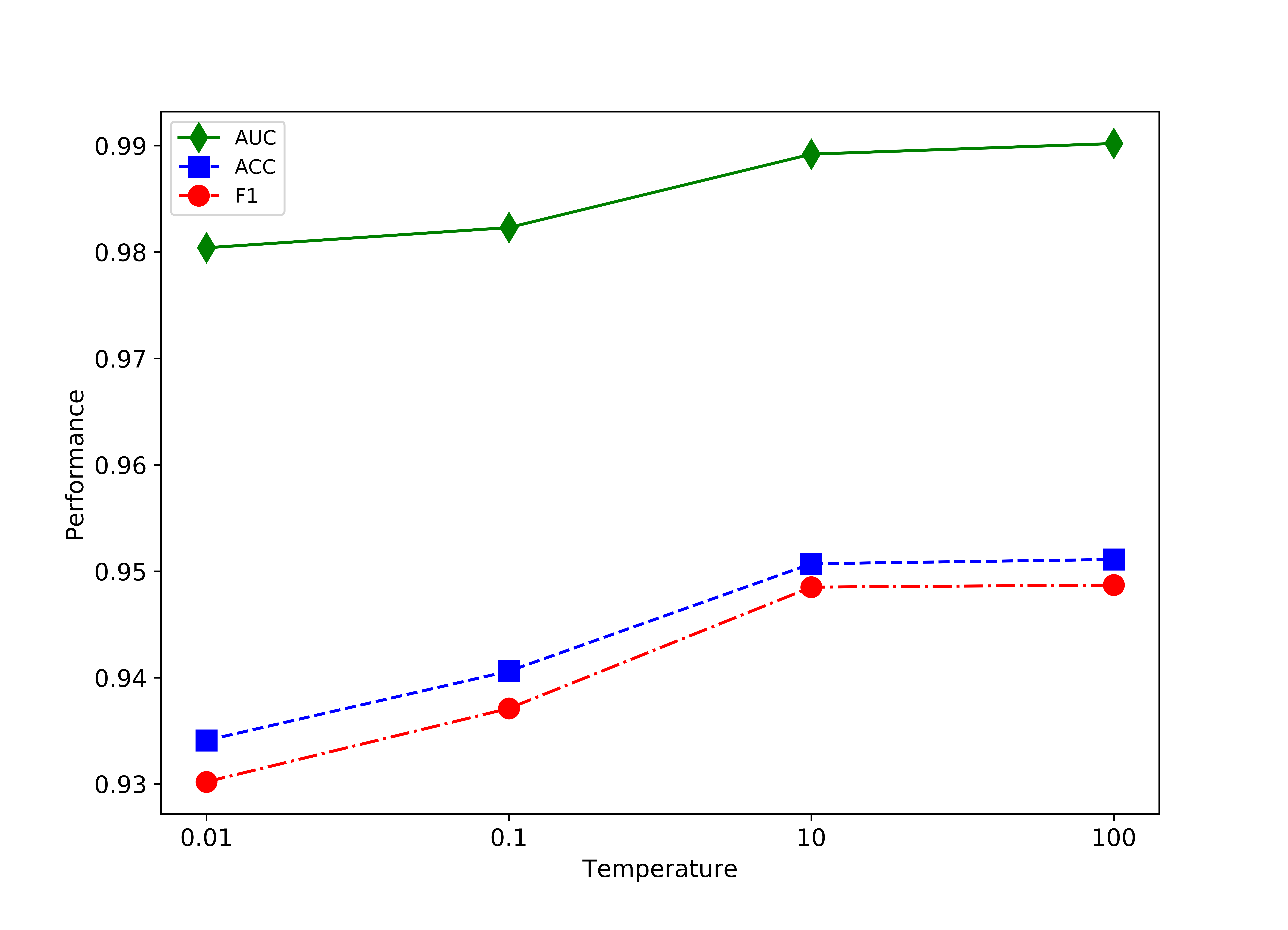}
    \caption{The performance (AUC, ACC, F1) of our LIFA with different Softmax temperatures of 0.01, 0.1, 10, and 100 on \textbf{AIVIVN} dataset.}
    \label{fig:temperature_aivivn}
\end{figure}
\begin{figure}[t]
    \centering
    \includegraphics[width=0.99\textwidth]{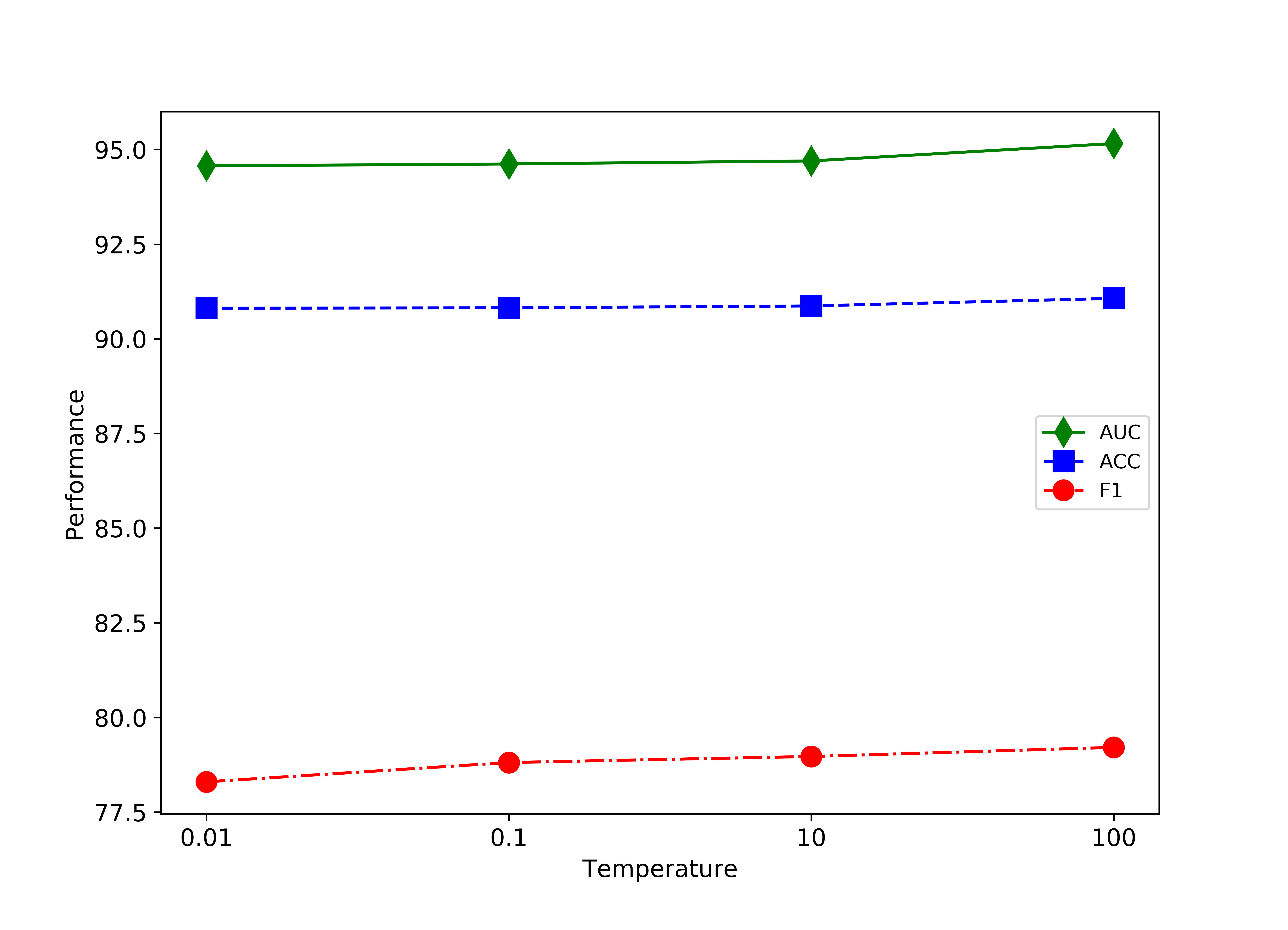}
    \caption{The performance (AUC, ACC, F1)  of our LIFA with different Softmax temperatures of 0.01, 0.1, 10, and 100 on \textbf{AISIA-VN-Review-S} dataset.}
    \label{fig:temperature_tiki}
\end{figure}

In this section, we explore the effect of prior knowledge sharing structure between experts and how it affects the final performance of LIFA. Our LIFA presents a simple, yet effective way for users to enforce certain behaviours among experts via a single temperate hyper-parameter in the softmax gating layer: experts could cooperate together (LIFA-COOP, high temperature) or compete against one another (LIFA-WTA, low temperature). By increasing the temperature to infinity, LIFA-COOP forces the experts to cooperate with one another and their predictions contribute equally to the final predictions. In contrast, LIFA-WTA lowers the temperature towards zero, which allocates all the weights to one experts who has the best performance. Since each expert's gradient is multiplied by its weight, experts making wrong decision will have their weights lowered towards zero, which vanishes the gradient signals. As a result, LIFA-WTA will aggressively selects a few experts who contribute the most to the correct predictions.

To verify such behaviours, we train several  LIFA-COOP and LIFA-WTA models with different temperature values and report the weight distribution $[\bm \alpha_{0}, \bm \alpha_{1}, ..., \bm \alpha_{n}]$ in Figures~\ref{fig:weight_distribution_aivivn} and \ref{fig:weight_distribution_tiki}. On both datasets, we observe that the weight distributions becomes smoother with higher temperature values, which supports our hypothesis. The results show that it is more beneficial for the experts to cooperate rather than to compete with one another. From the weight distributions, one can see that LIFA-WTA essentially performs model selection to choose the best performing experts. As a result, LIFA-WTA does not encourage knowledge sharing among experts and thus, performs poorly compared to LIFA-COOP. This experiment's results shed light on the success of LIFA-SIGMOID and LIFA-COOP by showing that encourage cooperation is more beneficial than model selection for transfer learning.

\section{Conclusion}
\label{sec:conclusion}
In this work, we studied sentiment classification with a focus on the Vietnamese language. We explored the potentials and limitations of the existing approaches and showed that it is beneficial to take advantage of multiple pre-trained models for transfer learning. This observation motivated us to propose LIFA, an efficient framework to learn a unified embedding from several pre-trained models (experts) and could perform better than its components. We further proposed a simple technique to enforce certain prior structures to such experts, resulting in two more LIFA variants that encouraged the experts to either cooperate or compete with one another. Moreover, we also constructed the {AISIA-VN-Review-F} dataset, which is the first large-scale sentiment classification database for the Vietnamese language. Through extensive experiments on several benchmarks, we demonstrated the efficacy of LIFA compared to existing techniques and comprehensively studied the benefits and drawbacks of its variants. We firmly believe our work will greatly contribute to the Vietnamese NLP research community. Finally, we will publish our codes and datasets used in this work upon acceptance.

\section*{Acknowledgments}
This research is funded by Vietnam National University Ho Chi Minh City (VNU-HCM) under grant number NCM2019-18-01. We want to thank the University of Science, Vietnam National University in Ho Chi Minh City, and  AISIA Research Lab in Vietnam for supporting us throughout this paper.

\bibliographystyle{elsarticle-num-names}
\bibliography{refs}

\end{document}